\documentclass[11pt,draftcls,onecolumn]{IEEEtran}

\usepackage[pdftex]{graphicx}
\graphicspath{{./Figures/}}
\DeclareGraphicsExtensions{.pdf,.jpeg,.jpg,.png}
\usepackage{amsmath,epsfig}
\usepackage{mathtools} 
\usepackage{amssymb}
\usepackage{subfig}
\usepackage{multirow}
\usepackage{setspace}
\usepackage[utf8]{inputenc} 
\usepackage[T1]{fontenc}    
\usepackage{url}            
\usepackage{booktabs}       
\usepackage{amsfonts}       
\usepackage{nicefrac}       
\usepackage{microtype}      
\usepackage{cite}			

\usepackage{mathtools}

\usepackage[usenames,dvipsnames]{color}

\newcommand{\lt}[1]{\textcolor{black}{#1}}

\begin{document}



\title{Graph signal processing for machine learning: A review and new perspectives}
\author{Xiaowen Dong*, Dorina Thanou*, Laura Toni, Michael Bronstein, and Pascal Frossard
\thanks{*Authors contributed equally.}}
\maketitle

The effective representation, processing, analysis, and visualization of large-scale structured data, especially those related to complex domains such as networks and graphs, are one of the key questions in modern machine learning. Graph signal processing (GSP), a vibrant branch of signal processing models and algorithms that aims at handling data supported on graphs, opens new paths of research to address this challenge. In this article, we review a few important contributions made by GSP concepts and tools, such as graph filters and transforms, to the development of novel machine learning algorithms. In particular, our discussion focuses on the following three aspects: exploiting data structure and relational priors, improving data and computational efficiency, and enhancing model interpretability. Furthermore, we provide new perspectives on future development of GSP techniques that may serve as a bridge between applied mathematics and signal processing on one side, and machine learning and network science on the other. Cross-fertilization across these different disciplines may help unlock the numerous challenges of complex data analysis in the modern age.

\section{Introduction}
\label{sec:intro}
We live in a connected society. Data collected from large-scale interactive systems, such as biological, social, and financial networks, become largely available. In parallel, the past few decades have seen a significant amount of interest in the machine learning community for network data processing and analysis \cite{Zhu05,Fortunato10,Nickel16}. 
Networks have an intrinsic structure that conveys very specific properties to data, e.g., interdependencies between data entities in the form of pairwise relationships. These properties are traditionally captured by mathematical representations such as graphs.

\begin{figure}[t]
      \centering
        \includegraphics[width=16cm]{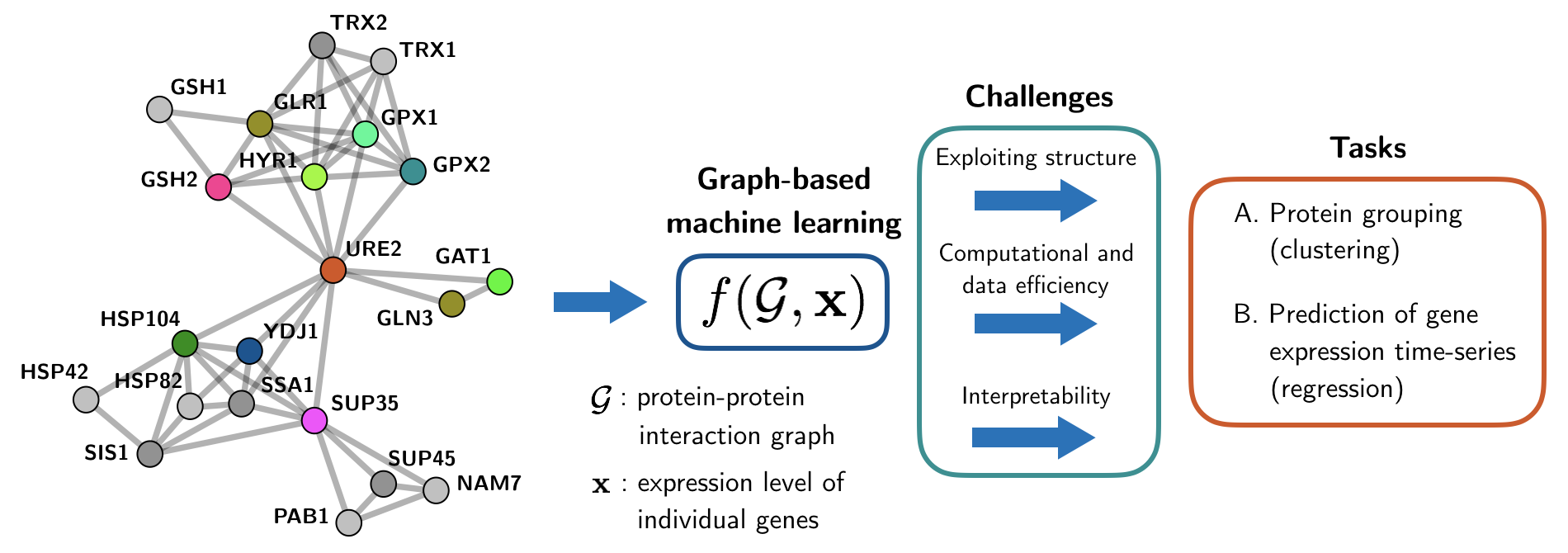}
        \caption{{Illustrative example of a typical graph-based machine learning task and the corresponding challenges. The input of the algorithm consists of i) a typical protein-protein interaction network captured by a graph, and ii) a signal on the graph (color-coded) that is an expression level of individual genes at any given time point. The output can be a classical machine learning task such as clustering of the proteins, or prediction of the gene expression over time. Figure inspired by Figure 1 in \cite{Szklarczyk19}.}}
        \label{fig:ppi}
\end{figure}

{In this context, new trends and challenges have been developing fast. Let us consider for example a network of protein-protein interactions and the expression level of individual genes at every point in time. Some typical tasks in network biology related to this type of data are {i) discovery of key genes (via protein grouping) affected by the infection and ii) prediction of how the host organism reacts (in terms of gene expression) to infections over time,} both of which can inform the best intervention strategies. These are two classical machine learning tasks, which involve learning with  graph-structured data (see Fig. \ref{fig:ppi} for an illustration). Furthermore, the rapid growth of gene and protein sequence data stretches the limit of graph-based algorithms, which need to be robust and stable against potential noise. Finally, the understanding and interpretability of the learning model {(such as the interpretation of the coefficients of the regression model)} are of crucial importance to clinical decision-making, for example the vaccine somministration.}

The quest for more efficient representation of signals collected in new structured observation domains has led to the recent development of graph signal processing (GSP) \cite{Shuman13,Sandryhaila13,Ortega18}. GSP is a fast-growing field at the intersection of applied mathematics and signal processing, which has seen a significant amount of research effort in the signal processing community at large. In these settings, data are modeled as signals defined on the node set of a weighted graph, which provide a natural representation that incorporates both the data (node attribute) and the underlying structure (edge attribute). By extending classical signal processing concepts, such as sampling, convolution, and frequency-domain filtering, to signals on graphs, GSP provides new ways of leveraging the intrinsic structure and geometry of the data in the analysis. This provides a generic framework and numerous appealing tools to address some of the new challenges encountered by current machine learning models and algorithms that handle network data. 

On the one hand, GSP provides new ways of exploiting data structure and relational priors from a signal processing perspective. This leads to both development of new machine learning models that handle graph-structured data, e.g., graph convolutional networks for representation learning \cite{bronstein2017geometric,Wu19}, and improvement in the efficiency and robustness of classical network algorithms, e.g., spectral clustering to detect network communities \cite{Tremblay19}. On the other hand, GSP provides a new perspective about the interpretability of popular machine learning models, e.g., the deep neural networks (DNNs), and gives additional insights on understanding real-world systems, e.g., by inferring the underlying network structure from observations \cite{Mateos19,Dong19}. Hence, GSP tools interestingly serve as a bridge that connects machine learning, signal processing, and network science towards solving important challenges in modern data analysis.

In this article, we review the application of GSP concepts and tools in developing novel as well as improving existing machine learning models.
{Our discussion revolves around three broad categories of GSP-based ideas: 1) regularization techniques to enforce smoothness of signals on graphs; 2) sampling techniques, filters, and transforms defined on graphs; {3) learning models inspired or interpretable by GSP concepts and tools.}}
We highlight their utilization for: 1) exploiting data structure and relational priors; 2) improving data efficiency, computational efficiency, and robustness; and 3) designing interpretable machine learning and artificial intelligence (AI) systems. 
We further provide a number of open challenges and new perspectives on the design of future GSP algorithms for machine learning and data analysis applications in general.

\section{GSP for exploiting data structure}
\label{sec:structure}

GSP typically permits to model, analyze, and process data supported on graphs. Consider a weighted and undirected\footnote{We assume an undirected graph for ease of discussion.} graph $\mathcal{G}=\{\mathcal{V},\mathcal{E}\}$ with the node set $\mathcal{V}$ of cardinality $N$ and edge set $\mathcal{E}$. A graph signal is defined as a function $\mathbf{x}: \mathcal{V} \rightarrow \mathbb{R}$ that assigns a scalar\footnote{In the more general case, the values on each node can be modeled as a vector as well.} value to each node. The main research effort in GSP is therefore concerned with the generalization of classical signal processing concepts and tools to graph signals. One key development is that the set of eigenvectors of the graph Laplacian matrix or a graph shift operator {(a generic matrix representation of the graph)} provides a notion of frequency on graphs and helps define the so-called graph Fourier transform (GFT). For example, consider the combinatorial graph Laplacian $\mathbf{L} = \mathbf{D} - \mathbf{W}$, where $\mathbf{W}$ is the weighted adjacency matrix of the graph and $\mathbf{D}$ is the degree matrix. The graph Laplacian $\mathbf{L}$ admits the eigencomposition $\mathbf{L} = \boldsymbol{\Phi} \mathbf{\Lambda} \boldsymbol{\Phi}^T$,
where $\boldsymbol{\Phi}$ is the eigenvector matrix that contains the eigenvectors as columns, and $\boldsymbol{\Lambda}$ is the eigenvalue matrix $\textbf{diag}(\lambda_1, \lambda_2, \cdots, \lambda_N)$ that contains the associated eigenvalues (conventionally sorted in an increasing order) along the diagonal. Due to the analogy between the continuous Laplace operator and the graph Laplacian $\mathbf{L}$, the operation $\langle \boldsymbol{\Phi}, \mathbf{x} \rangle = \boldsymbol{\Phi}^T \mathbf{x}$ can be thought of as a GFT for the graph signal $\mathbf{x}$, {with the $i$-th entry of $\boldsymbol{\Phi}^T \mathbf{x}$ being the Fourier coefficient that corresponds to the discrete frequency $\lambda_i$ \cite{Shuman13}\footnote{Similar transforms could be defined using a general graph shift operator \cite{Sandryhaila13,Ortega18}.}. 
This observation is useful in two aspects. First, the notion of frequency helps define a measure of smoothness for signals on graphs, and provides new interpretation of the theory of kernels and regularization on graphs \cite{Smola03}.}
Second, this enables the generalization of operations such as convolution and frequency-domain filtering for graph signals. GSP-based ideas thus provide new tools for exploiting data structure and relational priors. Since their inception, they have been applied in the context of a wide range of machine learning problems. {In this section, we discuss the application of GSP in supervised and unsupervised learning problems when the underlying graph is known; alternatively, it may be inferred from observations of different variables (see Section \ref{sec:graphlearning} for further discussion on this point).}

\subsection{Regression for multiple response variables}
Regression is one of the basic forms of supervised learning where, in its simplest setting, we are interested in learning a mapping from a set of inputs (or features) to a real-valued output (or response). Of particular interest is the generalized problem of multivariate regression, where we learn such a mapping for multiple response variables. Existing approaches in the literature usually rely on some types of modeling of the interdependencies between the response variables. For example, in a parametric model, such interdependencies are implicitly captured by the coefficients of a multivariate autoregressive process; in a nonparametric model, they may be explicitly modeled by the kernel function of a multi-output Gaussian process (GP). {In particular, in the case of separable kernels for multi-output GP, a positive semidefinite (PSD) matrix is incorporated into the kernel function to encode the interactions among the response variables (see Section 4 of \cite{Alvarez12}).}

GSP provides an interesting alternative to tackling such regression problem, by assuming that observations of the response variables form graph signals with certain properties (e.g., smoothness) on a predefined graph. For example, a globally smooth graph signal $\mathbf{y} \in \mathbb{R}^N$ can be obtained by solving a regularization problem on the graph and interpreted as outcome of the low-pass filtering of an input signal $\mathbf{x} \in \mathbb{R}^N$ (see Example 2 in \cite{Shuman13}):
\begin{equation}
\mathbf{y} = \mathbf{B} \mathbf{x} = (\mathbf{I}+ \alpha \mathbf{L})^{-1} \mathbf{x} = \boldsymbol{\Phi} (\mathbf{I}+ \alpha \mathbf{\Lambda})^{-1} \boldsymbol{\Phi}^T \mathbf{x},
\label{eq:gpg}
\end{equation}
where $\mathbf{I}$ is an identity matrix, and $\alpha$ is a hyperparameter. Notice that the filter matrix $\mathbf{B}$, which is a function of the graph Laplacian $\mathbf{L}$, encodes relationships between observations on different nodes. Based on this, the authors of \cite{Venkitaraman18} have gone on to propose a GP model on graphs, whose kernel function is reminiscent of that in a multi-output GP, {where $\mathbf{B}$ is used to construct the PSD matrix that relates observations of the response variables.}

The above example demonstrates the usefulness of GSP tools, e.g., {graph regularization and graph filters,} in designing models that capture interdependencies between multiple response variables in a regression setting. In particular, this enriches the multi-output GP literature by providing new ways of designing the kernel functions. {For example, in addition to the example in \cite{Venkitaraman18}, the recent work in \cite{Zhi20} has proposed to learn a graph filter that leads to the most suitable GP kernel for the observed graph data, hence improving the prediction performance.}

\subsection{Graph-based classification} 
Another basic form of supervised learning is classification, which is similar to regression but with a categorical response variable. Recent advances in deep learning, in particular convolutional neural networks (CNNs), have led to significant improvements in tasks such as image classification \cite{Krizhevsky12}. 
However, classical CNNs cannot be directly applied to classify signals that are supported on a graph structure, due to a lack of notion of shift and convolution in the irregular graph domain.

GSP provides a convenient tool to address such challenge. More specifically, by providing a notion of frequency and a GFT, convolution may be implicitly defined via the graph spectral domain. Consider a graph signal $\mathbf{x}$ and a convolutional kernel $\hat{g}(\lambda)$, which is defined directly in the graph spectral domain as a function applied to the Laplacian eigenvalues. We may define the convolution between $\mathbf{x}$ and $g$ as:
\begin{equation}
\mathbf{x} \ast g = \boldsymbol{\Phi} \hat{g}(\mathbf{\Lambda}) \boldsymbol{\Phi}^T \mathbf{x} = \hat{g}(\mathbf{L}) \mathbf{x}.
\label{eq:spec-conv}
\end{equation}
{In Eq.~(\ref{eq:spec-conv}), $\hat{g}(\mathbf{\Lambda}) = \textbf{diag}\big(\hat{g}(\lambda_1), \hat{g}(\lambda_2), \cdots, \hat{g}(\lambda_N)\big)$ thus the convolution operator $\hat{g}(\mathbf{L})$ can generally be interpreted as a graph filter, whose characteristic is determined by the form of $\hat{g}(\cdot)$ that can be learned in the context of a neural network.
This is exactly the idea behind the work in \cite{Bruna14}, where the authors have considered the problem of graph signal classification and proposed a spectral CNN whose architecture resembles that of a classical CNN, with the spatial filter replaced by a graph filter:}
\begin{equation}
\mathbf{x}_{l+1,q} = \sigma \big( \sum_{p=1}^{P} \boldsymbol{\Phi} \hat{g}_{\boldsymbol{\theta}_{q,p}}(\mathbf{\Lambda}) \boldsymbol{\Phi}^T \mathbf{x}_{l,p} \big).
\label{eq:spec-cnn}
\end{equation}
In Eq.~(\ref{eq:spec-cnn}), $\mathbf{X}_l \in \mathbb{R}^{N \times P} = [ \mathbf{x}_{l,1}, \mathbf{x}_{l,2}, \cdots, \mathbf{x}_{l,P} ]$ and $\mathbf{X}_{l+1} \in \mathbb{R}^{N \times Q} = [ \mathbf{x}_{l+1,1}, \mathbf{x}_{l+1,2}, \cdots, \mathbf{x}_{l+1,Q} ]$ are the feature maps at layer $l$ and $l+1$, respectively, $\hat{g}_{\boldsymbol{\theta}_{q,p}}(\mathbf{\Lambda}) = \textbf{diag}(\boldsymbol{\theta}_{q,p})$ where $\boldsymbol{\theta}_{q,p} \in \mathbb{R}^N$ are the learnable parameters, and $\sigma(\cdot)$ is a nonlinearity applied to the node-wise signal values. In this construction, node-domain convolution is implicitly carried out in the graph spectral domain by an element-wise multiplication between the Fourier coefficients of the signal, $\boldsymbol{\Phi}^T \mathbf{x}_{l,p}$, and the learnable parameter vector, $\boldsymbol{\theta}_{q,p}$, {before going back to the node domain.} This early work thus highlights the benefits of GSP tools in addressing an important challenge in the design of graph neural networks (GNNs) \cite{bronstein2017geometric,Wu19}.

\begin{figure}[t]
      \centering
        \includegraphics[width=14cm]{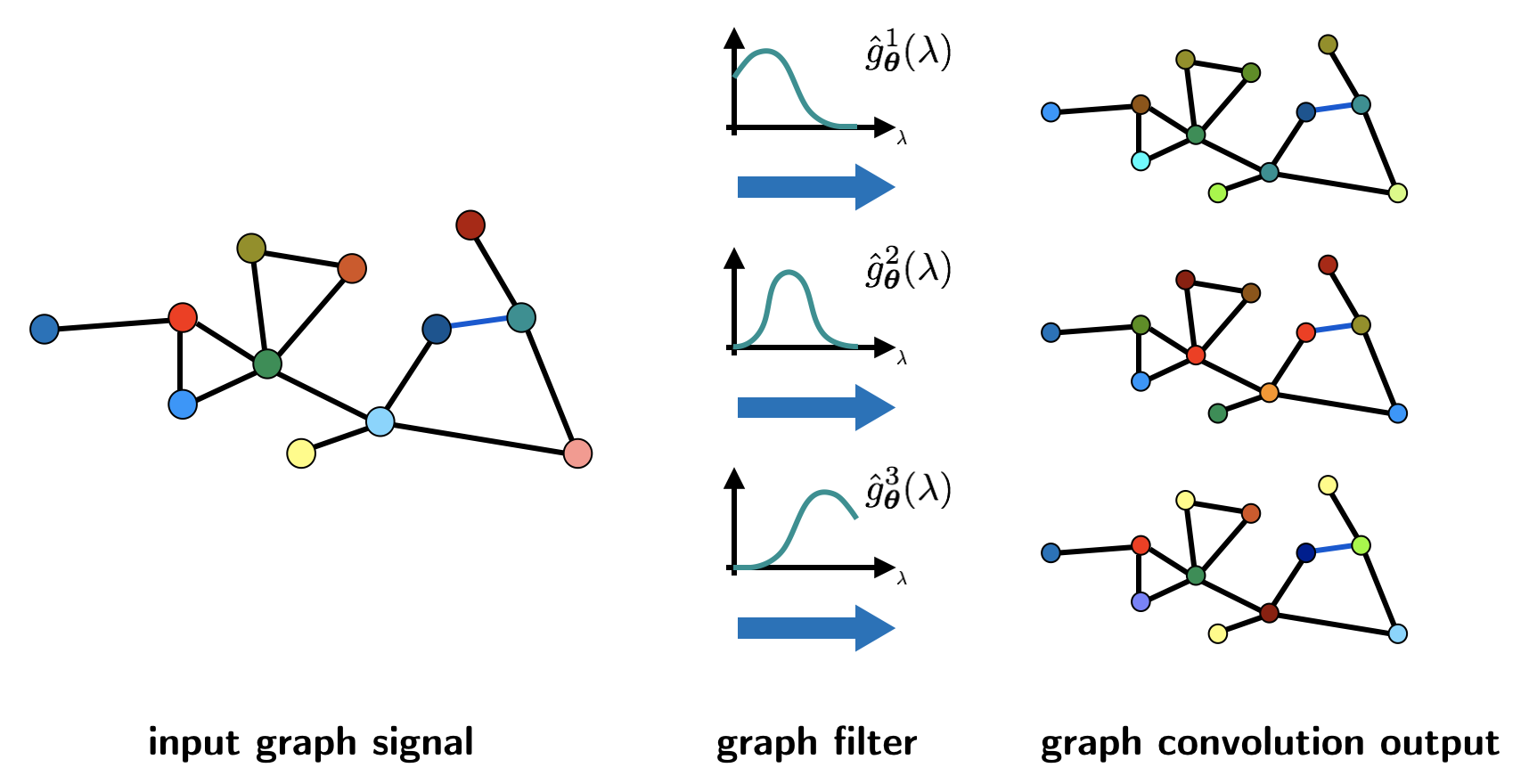}		
        \label{PPI}
        \caption{{Graph convolution via polynomial spectral graph filters proposed in the ChebNet architecture. An input graph signal is passed through three graph filters separately to produce output of the first graph convolutional layer (before nonlinear activation and graph pooling) of the ChebNet. Figure inspired by Figure 1 in \cite{defferrard2016convolutional}.}}
        \label{fig:chebynet}
\end{figure}

The work in \cite{Bruna14} is conceptually important, but it is limiting in a number of ways. In particular, the convolutional filter defined in the graph spectral domain is not necessarily localized in the spatial (node) domain, a property that is often desired in neural network design. {Furthermore, the necessity for computing the eigendecomposition (with a complexity of $\mathcal{O}(N^3)$) as well as the GFT via the matrix-vector multiplication (with a complexity of $\mathcal{O}(N^2)$) prevents this framework to be applied to large-scale graphs.} Motivated by these limitations, Defferrard et al. have proposed a graph CNN framework, also known as the ChebNet \cite{defferrard2016convolutional}. {The key idea in \cite{defferrard2016convolutional} is the design of the convolutional filter via a finite polynomial of the eigenvalues of the graph Laplacian:}
\begin{equation}
{\hat{g}_{\boldsymbol{\theta}}(\mathbf{\Lambda}) = \sum_{j=0}^K \theta_j \mathbf{\Lambda}^j~~~\text{thus}~~~\hat{g}_{\boldsymbol{\theta}}(\mathbf{L}) = \sum_{j=0}^K \theta_j \mathbf{L}^j,}
\label{eq:poly}
\end{equation}
{where $\boldsymbol{\theta} = [\theta_0, \cdots, \theta_K] \in \mathbb{R}^{K+1}$ are the learnable parameters. It has been pointed out in \cite{defferrard2016convolutional} that, due to the property of the $K$-degree polynomial of the Laplacian,} the convolutional filter is guaranteed to be localized within the $K$-hop neighborhood of each node. Furthermore, the convolution can be done by matrix-vector multiplication using the powers of the graph Laplacian; no explicit computation of the eigendecomposition or the GFT is required. Finally, to improve the stability of the convolution operation, the authors of \cite{defferrard2016convolutional} consider a scaled version of the graph Laplacian, $\tilde{\mathbf{L}}$, and a Chebyshev approximation to $\hat{g}_{\boldsymbol{\theta}}(\tilde{\mathbf{L}})$, hence the name of the approach. ChebNet represents an important advance in GNN designs, {and its development is based on the core idea of graph convolution via polynomial spectral graph filters in Eq.~(\ref{eq:poly}). An illustration of the graph convolutional layer of the ChebNet is shown in Fig. \ref{fig:chebynet}.}

The works in \cite{Bruna14,defferrard2016convolutional} consider supervised learning (classification) of graph signals, an exact analogy to image classification via traditional CNNs. Another popular graph-based classification problem is semi-supervised node classification, where one is interested in inferring the label for a set of testing nodes given labels on a set of training nodes, as well as the connectivity and any additional features for the full set of nodes. Although there exists an extensive literature on semi-supervised learning on graphs \cite{Zhu05,Chapelle06}, 
the perspective of neural network based approaches provides renewed interest in the problem from the machine learning community. A major development along this line is the graph convolutional network (GCN) proposed in \cite{Kipf17}, which can be thought of as a special case of the ChebNet design by letting $K=1$. After a series of simplifications, one can express the convolutional layer of the GCN as:
\begin{equation}
\hat{g}_{\theta}(\mathbf{L}) = \theta \mathbf{\tilde{D}}^{-1/2} \mathbf{\tilde{W}} \mathbf{\tilde{D}}^{-1/2}.
\label{eq:gcn}
\end{equation}
In Eq.~(\ref{eq:gcn}), {$\mathbf{\tilde{D}}$ is a diagonal matrix with the $i$-th entry of the diagonal being $\mathbf{\tilde{D}}_{ii} = \sum_{j} \tilde{W}_{ij}$, where $\tilde{W}_{ij}$ is the $ij$-th entry of the matrix $\mathbf{\tilde{W}} = \mathbf{W} + \mathbf{I}$,} and $\theta$ is the single learnable parameter for each filter. The GCN can be considered as a neural network that updates node features for node classification, hence does not require a graph pooling layer as in the case of \cite{Bruna14,defferrard2016convolutional}. The update process is a simple one-hop neighborhood averaging, {as implied by the choice of $K=1$ for the convolutional filter and reflected by the form of normalized adjacency matrix in Eq.~(\ref{eq:gcn}).} This also demonstrates the interesting point that, although GNN designs such as the ChebNet and the GCN might be motivated from a graph spectral viewpoint, they may be implemented in the spatial (node) domain directly. The work in \cite{Kipf17} has demonstrated that the GCN significantly outperforms a number of baselines in graph-based semi-supervised learning tasks, {and it has since sparked a major interest in developing GNN models from the machine learning community.}

In summary, GSP tools, and especially the graph filters, have played an important role in some early designs of GNN architectures. {Furthermore, the recent study in \cite{Fu20} has shown that several popular GNNs can be interpreted as implementing denoising and/or smoothing of graph signals. GNNs, especially those designed from a spectral perspective, can therefore be thought of as learning models inspired or interpretable by GSP concepts and tools.} They have not only contributed to graph-based classification but also other aspects of machine learning, as we explain in the following sections.

\subsection{Graph-based clustering and dimensionality reduction}
Unsupervised learning is another major paradigm in machine learning, where clustering and dimensionality reduction are two main problems of interest. Graph-based clustering, in its simplest form, aims to partition the node set into mutually exclusive clusters using a similarity graph. This problem has attracted major interest in the network science and machine learning communities \cite{Fortunato10}. Spectral clustering \cite{Luxburg07}, 
for example, is one of the most popular solutions that has been introduced in the last two decades.

Although graph-based clustering does not typically involve node attributes (i.e., graph signals), concepts that are provided by GSP have found their application in a number of related problems. One example is multi-scale clustering, where one is interested in finding a series of clustering results each of which reflects the grouping of nodes at a particular scale. While classical solutions such as the work in \cite{Lambiotte10} are usually based on generalization of concepts such as modularity \cite{Newman06} 
to the multi-scale scenario, multi-scale transforms in signal processing, such as the wavelet transform, provide a natural alternative. For example, the work in \cite{Tremblay14} has utilized the spectral graph wavelets \cite{Hammond11}, a generalization of wavelets to the graph domain, to address the problem. {Specifically, the graph wavelet $\boldsymbol{\psi}_{s,v}$ centered around node $v$ at a particular scale $s$ is defined as:
\begin{equation}
\boldsymbol{\psi}_{s,v} = \boldsymbol{\Phi} \hat{g}(s \mathbf{\Lambda}) \boldsymbol{\Phi}^T \boldsymbol{\delta}_v = \hat{g}(s \mathbf{L}) \boldsymbol{\delta}_v,
\label{eq:wave}
\end{equation}
where $\hat{g}(s \mathbf{\Lambda}) = \textbf{diag}\big(\hat{g}(s \lambda_1), \hat{g}(s \lambda_2), \cdots, \hat{g}(s \lambda_N)\big)$, and $\boldsymbol{\delta}_v$ is a delta impulse that has value 1 at node $v$ and 0 elsewhere. It is clear from Eq.~(\ref{eq:wave}) that the graph wavelet $\boldsymbol{\psi}_{s,v}$ can be interpreted as outcome of a scaled (in the graph spectral domain) graph filter given the delta impulse as input. It provides an ``egocentered view'' of the graph seen from node $v$, and correlation between such views therefore provides a similarity between the nodes at scale $s$.} A simple hierarchical clustering is then applied to the similarity matrix at different scales to obtain multi-scale clustering. Another example of GSP approach for clustering is proposed in \cite{Tremblay16}, {again based on the graph filters,} which we explain in more detail in Section \ref{sec:robustness} from an efficiency viewpoint.

Unsupervised learning can also take the form of dimensionality reduction problems. In the context of graph-structured data, this usually means learning a low-dimensional representation for the data entities in the presence of a graph that captures the pairwise similarity between them. Classical solutions to dimensionality reduction, such as matrix factorization and autoencoders, have recently been combined with advances in GNNs to provide novel solutions. For example, the work in \cite{Kipf16} has proposed an architecture based on the GCN to train a graph autoencoder for learning a low-dimensional representation of the data. Another example is the work in \cite{monti2017geometric}, where the authors have proposed an architecture that combines the ChebNet with a recurrent neural network architecture for matrix completion and factorization. {Both the GCN and ChebNet are based on the idea of polynomial spectral graph filters, which demonstrates how such GSP-inspired models can benefit learning algorithms for dimensionality reduction problems.}

\section{GSP for improving efficiency and robustness}
\label{sec:robustness}

{Although GSP is a framework originally developed for incorporating the graph structure into data analysis, since recently, many classical machine learning algorithms have also benefited from exploiting typical GSP tools. Significant improvement in terms of  robustness with respect to scarse and noisy training data, or adversarial examples, has been shown in tasks such as (semi-)supervised learning, few-shot learning, zero-shot learning, and multi-task learning. }
Furthermore, large-scale data analysis and statistical learning can benefit from typical GSP operations such as sampling and filtering to reduce the computational complexity and time of graph-based learning. We discuss some of these aspects in the following sections.

\subsection{Improvement on data efficiency and robustness}
Graph regularization has been introduced in many learning architectures, including DNNs, in order to improve the performance of supervised and semi-supervised learning tasks. In semi-supervised settings, the loss term has been augmented by a graph-based penalization term that biases the network towards learning similar representations for neighboring nodes on a graph that contains as nodes both labeled and unlabeled data \cite{Ye_2019}. 
In supervised learning problems, regularization technique that encourages smoothness of the {label signal on a similarity graph constructed from the data} \cite{Bontonou19} has been shown to improve the robustness of deep learning architectures by maximizing the distance between outputs for different classes. {Similar regularization idea can be applied to intermediate representations in an DNN; for example,} the regularizer proposed in \cite{Lassance19} is based on constructing a series of graphs, one for each layer of the DNN architecture, where each graph captures the similarity between all training examples given their representations at that layer. It aims at achieving robustness by penalizing large changes in the smoothness of class indicator vectors from one layer to the next. {Both {\cite{Bontonou19} and \cite{Lassance19}} demonstrate the benefit of using regularization techniques on graphs to introduce inductive bias towards more effective learning.}

{Few-shot learning is another domain of machine learning where GSP-inspired models, in particular the GNNs, have been shown to improve generalization to novel tasks using only a few examples from the unknown classes.}
{For instance, GCNs have been introduced in classical learning architectures: in \cite{Shi_2019}, seen and unknown classes are represented by prototypes that are learned jointly; in \cite{garcia2018fewshot}, a graph is constructed from images in all classes and it is eventually given as an input to a GCN.
On the other hand, zero-shot learning, i.e., when samples of the unknown classes do not even exist in the training set, has also benefited from GSP tools such as transforms defined on graphs.} As an example, the authors of \cite{Deutsch_2019} have introduced the isoperimetric loss as a regularization criterion for learning the map from a visual representation to a semantic embedding, in a zero-shot learning scenario. {Sample embeddings are modeled as nodes of a graph, whose connectivity is learned in the training phase. Smoothness on the graph is then achieved through the isoperimetric loss, which is approximated using polynomial functions that correspond to the spectral graph wavelets \cite{Hammond11}.}

\begin{figure}[t]
      \centering
        \includegraphics[width=12cm]{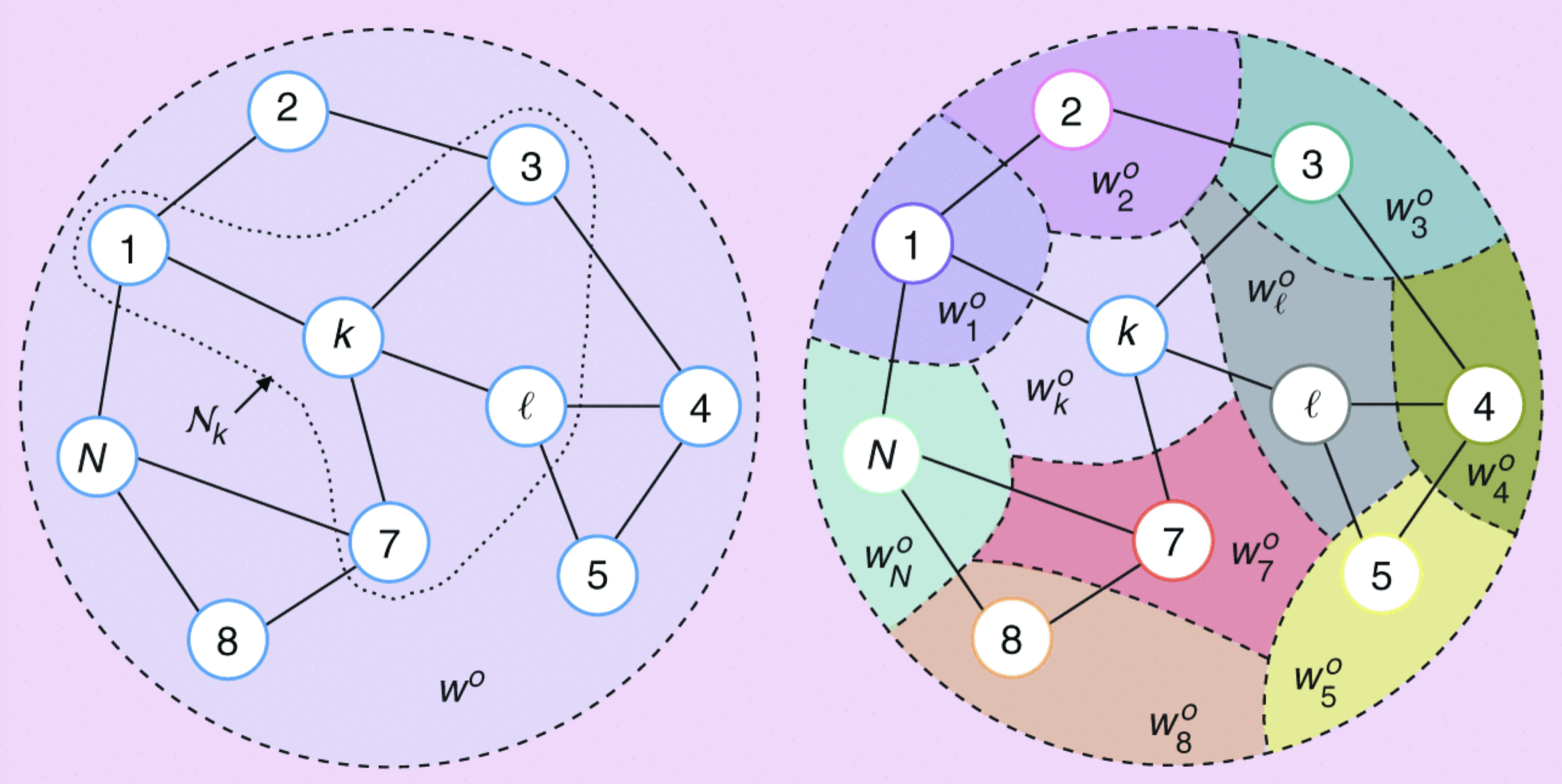}
        \caption{Illustrative example of a single-task network (left) and a multi-task network (right). {Nodes represent individual tasks each of which is to infer a parameter vector $\boldsymbol{w}^o$ that optimizes a differentiable cost function. In a single-task network, all tasks optimise for the same parameter vector $\boldsymbol{w}^o$, while in a multi-task network, they optimise for distinct, though related, parameter vectors \{$\boldsymbol{w}_\cdot^o$\}.} Figure from \cite{Nassif20} with permission.}
        \label{fig:multitask}
\end{figure}

Generalization across tasks is also key to multi-task learning, a field of research that has been recently addressed with GSP in \cite{Nassif20}. In this setting, the graph captures the correlation between multiple tasks, with each task being a node on the graph (see Fig. \ref{fig:multitask} for an illustration). The overall goal is for each task to infer the parameter vector that optimizes a differentiable cost function. Looking at the cost function as a signal on the graph, the knowledge from other tasks can be exploited by imposing a regularizer on the task graph, promoting the existing relationships between the tasks. This work shows how the structure between tasks can be inferred and exploited in learning-to-learn (or meta-learning) settings, generalizing the learning process across tasks.

\subsection{Robustness against topological noise}
In addition to robustness in terms of data samples, another natural consideration is the stability and robustness of graph-based models against topological noise. This usually means perturbation in the graph topology, which may be due to noise observed in real-world graphs or inaccuracy in the inference of the graph topology if one is not readily available. Stability and robustness against such perturbation is key to generalization capability of graph-based machine learning models.

It is well known that Laplacian eigenvectors are not stable against perturbation in the graph topology. 
Stability of GSP-based tools, especially the Laplacian-based spectral graph filters and spectral domain designs of GNN architectures, has come under scrutiny. A few recent results have however shown that spectral graph filters are, under certain conditions, indeed stable. 
For example, the work in \cite{levie2019transferability} has shown that spectral graph filters in the Cayley smoothness space are stable under absolute perturbation in the graph Laplacian, in the sense that the change in the filter output is linear with respect to the amount of perturbation. They have further analyzed the transferability of spectral domain designs of GNNs between graphs that discretize the same underlying metric space. The work in \cite{Gama19} has analyzed stability of graph filters with respect to both absolute and relative graph perturbation, and used the results to bound the change in the output of a GNN architecture as a function of properties of the perturbation (i.e., eigenvector misalignment and upper bound for the norm of the perturbation matrix), filter characteristic (i.e., integral Lipschitz constant of the filter), and design of the GNN architecture (i.e., number of filters and layers). 
Similarly, the recent work in \cite{Kenlay20} has shown the stability of spectral graph filters that are based on polynomials of the graph Laplacian.

In addition to the stability results above, there exist several other studies on the impact of perturbation or uncertainty in the graph topology on GSP and network analysis in general. For example, 
the work in \cite{Isufi17} has analyzed polynomial spectral graph filters with random graph realizations, while the work in \cite{Ceci18b} has focused on changes in the Laplacian eigenvalues and eigenvectors. The work in \cite{Segarra18}, on the other hand, has looked at stability and continuity of centrality measures in weighted graphs.
{All these examples are important attempts in understanding the robustness of network analysis, GSP operations, and GSP-inspired models against topological noise.}

\subsection{Improvement on computational efficiency}
The benefit of graph priors in machine learning tools in terms of data efficiency, robustness, and generalization has been empirically and theoretically supported. Nevertheless, these learning frameworks usually scale poorly with the number of graph nodes, resulting in high computational complexity in large-scale problems. Hence, there exists a strand of research aimed at adopting GSP tools to reduce the computational complexity of classical learning frameworks, such as spectral clustering and spectral sparsification. 

Spectral clustering is a typical example of highly useful yet expensive frameworks. The three main steps of spectral clustering are: 1) graph construction for the $N$ entities to cluster; 2) computation of the first $k$ eigenvectors of the graph Laplacian to identify a feature vector (embedding) for each entity; and 3) application of the $k$-means algorithm to these feature vectors. Each of these three steps becomes computationally intensive for large number of nodes $N$ and/or target number of clusters $k$. Fundamental GSP techniques, such as graph filtering and random signal sampling, have been proposed to simplify the last two steps of spectral clustering, leading to a compressive spectral clustering algorithm \cite{Tremblay16}. In particular, the computation of the eigenvectors is bypassed by filtering random signals on the graph. The $k$-means step is then performed on a small number (instead of $N$) of randomly selected feature vectors, exploiting the theory of random sampling of bandlimited graph signals. We refer the reader to \cite{Tremblay19} for an overview of the recent works aimed at reducing the computational complexity of spectral clustering.

{More broadly, analysis and processing of data that reside on large-scale graphs can be simplified by factorizing the complex graph structure into the product of smaller graphs. This is the idea behind the work in \cite{Sandryhaila14}, {which adopts product graph as a model for large graphs,} and defines signal processing operations, such as the GFT and filtering, in a more efficient manner.
For example, a Cartesian product graph can be used to modularize a regular grid image in order to gain in terms of computational efficiency in the computation of eigendecomposition for the GFT. This enables efficient implementation of GSP analysis on large graphs via parallelization and vectorization.}

\section{GSP for enhancing model interpretability}
\label{sec:interpretability}

{A third domain that has recently benefited from the advances in the GSP literature is the interpretation of complex systems and well as classical machine learning algorithms.} In this section, we focus our discussion on two aspects: 1) inferring hidden structure from data that leads to a better understanding of a complex system; and 2) explaining the outcome of classical ``black-box'' learning architectures.

\subsection{Inference of hidden relational structure for interpretability}
\label{sec:graphlearning}
{Analysis of complex systems typically involves large sets of data whose underlying structure is often unknown. Such structure can be estimated from the data in order to permit effective understanding or visualization of the system behavior. For example, in neuroscience, inferring a structure in the form of the connectivity between brain regions from activation signals may lead to a better understanding of the functionality of the brain.}

The problem of structure or topology inference has been studied from classical viewpoints in statistics (e.g., sparse inverse covariance estimation \cite{Banerjee08,Friedman08}) or physics (e.g., cascade models \cite{GomezRodriguez_2010,Myers2010}). More recently, the problem has been addressed from a GSP perspective, where the structure is modeled as a graph that is to be inferred by exploiting the interplay between the graph and the signals observed on the nodes. 
{GSP-based graph-learning frameworks therefore have the unique advantage of enforcing certain desirable representations of the signals via frequency-domain analysis and filtering operations on graphs.}
For example, models based on assumptions such as the smoothness \cite{Dong16,Kalofolias16,Egilmez17,Chepuri17} 
or the diffusion \cite{Thanou17, Pasdeloup18, Segarra17a} 
of the graph signals have been successfully applied to identify weather patterns, brain connectivity, mobility patterns, and 3D point cloud structure \cite{Chen2020,wang2019dynamic}. 
Due to space limitation, we do not provide details of these methods in this paper and refer the interested readers to two overview papers on the topic \cite{Mateos19,Dong19}.

Furthermore, topology inference has been introduced in the context of DNNs/GNNs in order to infer hidden, and more interpretable, structures. For example, the authors in \cite{Tong_2020} have proposed to impose a graph-based regularization on features of each layer of the DNN, based on the signal smoothness, by simultaneously learning the feature space graph with nodes being the neurons. Specifically, features associated with the neurons in the intermediate layers are constrained to take into account the structure of a graph through a smoothness promoting regularizer. Such an approach is shown to increase the interpretability of the DNN layers.
Another example is the graph attention network (GAT) \cite{Velickovic18} which is a form of GNNs that adaptively updates edge weights in a data-driven manner, and can thus be used to interpret importance of neighbors in feature aggregation. {Interestingly, the recent work in \cite{Fu20} has shown that the GAT itself can be interpreted as implementing denoising of edge weights in a graph signal denoising problem.}

\subsection{A posteriori interpretation of learning architectures}
\begin{figure}
\centering
\includegraphics[width=16cm]{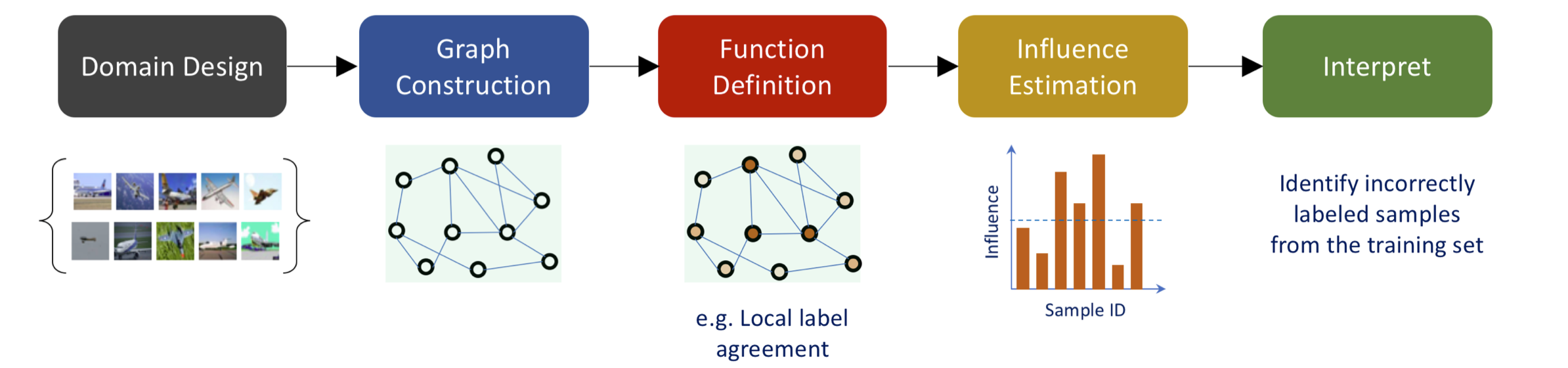}
\caption{Illustrative example of MARGIN. Figure from \cite{Anirudh17} with persimission.}\label{MARGIN}
\end{figure}

A first attempt to exploit GSP tools in order to better understand complex learning architectures is the framework proposed in \cite{Anirudh17} called MARGIN. {It provides a generic way to perform several meta-learning tasks, by allowing the user to incorporate rich semantic information. More specifically, for each task, a graph whose nodes represent entities of interest is constructed, and a hypothesis is defined as a function or signal on that graph.} {For example, if the task is to determine faulty labels in a data set with corrupted labels, the domain is the set of samples, and the hypothesis becomes a local label agreement signal that measures how many neighbors have the same label as the current node.} The central idea of MARGIN is then to use GSP to identify fluctuations of the signal on the graph, which serve as explanations. In particular, high-pass graph filtering reveals which nodes can maximally describe the variations in the label agreement signal. The node influence scores are computed as the magnitude of the filtered signal value at that node. The influence score is eventually translated into an interpretable explanation. A summary of the different steps of the approach is illustrated in Fig. \ref{MARGIN}. MARGIN is an interesting attempt based on GSP tools; however, one clear disadvantage of this method is the fact that if the domain is chosen to be the entire data set, the graph construction and the filtering operators become computationally intensive as the size of the data set increases.

Another example for improving interpretability is the work in \cite{Gripon18}, which uses GSP to monitor the intermediate representations obtained in a DNN. Preliminary finding suggests that smoothness of the label signal on a $k$-nearest neighbor graph, which is obtained using normalized Euclidean distance of features in each layer, is a good measure of separation of classes in these intermediate representations. {This thus constitutes a first step towards exploiting the frequency behavior of the label signal on some well-defined graphs across the layers in order to provide meaningful interpretations of the DNN layers.} {Besides global smoothness, other spectral properties of graph signals such as local smoothness or sparsity, as well as graph metrics such as centrality and modularity, could be worth investigating towards the same objective.}

\section{Summary, open challenges, and new perspectives}
\label{sec:conclusion}

Handling complex and dynamic structures within large-scale data sets presents significant challenges in modern data analytics and machine learning tasks. The novel signal processing perspective reviewed in this article  has therefore both theoretical and practical significance. 
On the one hand, GSP provides a new theoretical framework for incorporating complex structure within the data, and contributes to the development of new data analysis and machine learning frameworks and systems. Furthermore, GSP tools have been utilized to improve existing machine learning models, mainly in terms of data and computational efficiency, robustness against noise in data and graph structure, and model interpretability. Some of the representative contributions discussed in this paper are summarized in Table~\ref{tab:summary}. On the other hand, machine learning problems such as classification, clustering, and reinforcement learning, as well as their application in real-world scenarios, also provide important motivations for new theoretical development in GSP.
For these reasons, there has been an increasing amount of interest in the GSP community in engaging with machine learning applications. We provide below a few open challenges and new perspectives on future investigation.

\begin{table}[t]
\centering
\caption{{A summary of contributions of GSP concepts and tools to machine learning models and tasks.}}
\label{tab:summary}
\scalebox{0.8}{
\begin{tabular}{|l|c|c|c|}
\hline
\multicolumn{1}{|c|}{} & \begin{tabular}[c]{@{}c@{}}Graph-based smoothness \\ and regularization\end{tabular} & \begin{tabular}[c]{@{}c@{}}Graph sampling, filters, \\ and transforms \end{tabular} & \begin{tabular}[c]{@{}c@{}}GSP-inspired/interpretable \\ models\end{tabular} \\ \hline
\begin{tabular}[c]{@{}l@{}}Exploiting data \\ structure\end{tabular} & \begin{tabular}[c]{@{}c@{}}{GP \& Kernels on graphs \cite{Venkitaraman18,Zhi20}} \end{tabular} & \begin{tabular}[c]{@{}c@{}}{GP \& Kernels on graphs \cite{Venkitaraman18,Zhi20}}\\ {Multi-scale clustering \cite{Tremblay14}}\end{tabular} & \begin{tabular}[c]{@{}c@{}}Spectral GNNs \\ \cite{Bruna14,defferrard2016convolutional,Kipf17,Kipf16,monti2017geometric} \end{tabular} \\ \hline
\begin{tabular}[c]{@{}l@{}}Improving efficiency \\ and robustness\end{tabular} & \begin{tabular}[c]{@{}c@{}}{Improving DNNs \cite{Ye_2019,Bontonou19,Lassance19}}\\ {Multi-task learning \cite{Nassif20}} \end{tabular} & \begin{tabular}[c]{@{}c@{}}{Zero-shot learning \cite{Deutsch_2019}}\\ {Stability of filters \& GNNs \cite{levie2019transferability,Gama19,Kenlay20}}\\ {Spectral clustering \cite{Tremblay16,Tremblay19}}\\ {Large-scale data processing \cite{Sandryhaila14}}\end{tabular} & \begin{tabular}[c]{@{}c@{}}Few-shot learning \\ \cite{Shi_2019,garcia2018fewshot}\end{tabular} \\ \hline
\begin{tabular}[c]{@{}l@{}}Enhancing model \\ interpretability\end{tabular} & \begin{tabular}[c]{@{}c@{}}{Interpreting DNNs \cite{Tong_2020,Gripon18}} \end{tabular} & \begin{tabular}[c]{@{}c@{}}{Topology inference \cite{Mateos19,Dong19}}\\ {Interpreting DNNs \cite{Anirudh17}}\end{tabular} & \begin{tabular}[c]{@{}c@{}}{GAT \cite{Velickovic18}} \end{tabular} \\ \hline
\end{tabular}}
\end{table}

\subsection{GSP and probabilistic modeling}
One important direction is the connection between GSP, machine learning, and probabilistic modeling of the graph topology. Most existing GSP and machine learning frameworks and tools follow a deterministic setting, where a fixed graph topology is predefined or inferred from data. Real-world networks, however, are often noisy, incomplete, or evolving constantly over time. It is therefore important to take into account the uncertainty in the graph topology in the analysis. First, it would be important to study analytical models for noise in the graph topology and its impact on GSP analysis tasks. For example, the work in \cite{Miettinen19} has used the Erdös-Rényi graph, a well-known random graph model, as model for topological noise and analyzed its impact on the filtering operation and independent component analysis (ICA) of graph signals. Second, it would be interesting to take a Bayesian viewpoint to model the uncertainty in the graph topology. For example, the observed graph may be considered as an instance generated from a parametric model, and the posterior of the parameters may be computed so that new instances of the graph may be generated. This is exactly the idea behind the work in \cite{Zhang19}, which has adopted a mixed membership stochastic block model for the graph and proposed a Bayesian GCN framework. Both are promising future directions to combine GSP and machine learning in a coherent probabilistic framework.

\begin{figure}[t]
    \centering
\includegraphics[width=14cm]{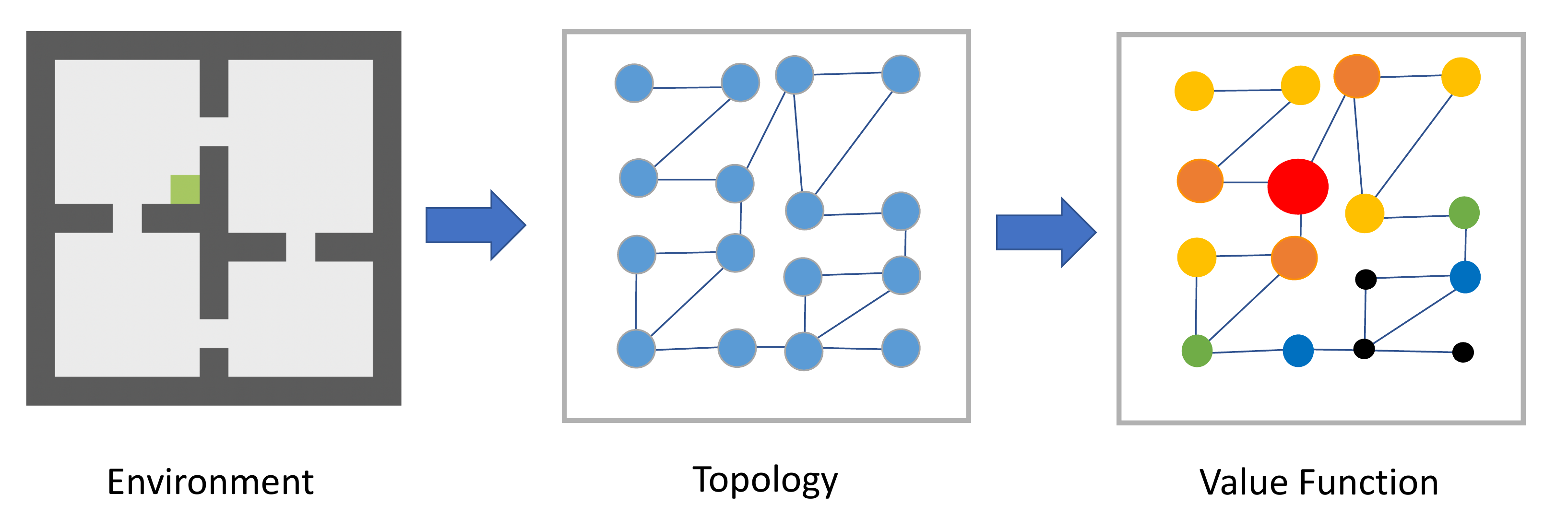}
\caption{Maze as an RL problem: an environment is represented by its multiple states -- ($x,y$) coordinates in the space (left). \lt{We  quantize the maze space into four  positions (therefore states) per room.}  States are identified as nodes of a graph, and state transition probabilities as edge weights (middle). The state value function can then be modeled as a signal on the graph (right), where both the color and size of the nodes reflect the signal values.}
\label{fig:graphs_DMS}
\end{figure}

\subsection{GSP and decision-making strategy}
{It is interesting to notice the link recently established between GSP and decision-making strategies (DMSs) under uncertainty, in which an agent needs to accomplish a task (maximize a cumulative or long-term reward) without any prior information. The environment (state space in reinforcement learning \cite{Mahadevan:2007} or context in bandit problems \cite{Toni:18}) can be modeled as a graph and instantaneous or long-term rewards can be seen as a signal on the graph (see Fig. \ref{fig:graphs_DMS} for an illustration). GSP tools can then be used to provide a sparse representation of the signal, and reduce the learning of the agent to a low-dimensional space. This improves the learning rate in DMS problems, which otherwise does not scale with the dimension of the state space, leading to highly-suboptimal decisions in large-scale problems. Many open challenges, however, still remain overlooked. First, the graph in these problems is usually constructed and not adaptively inferred, a scenario where GSP-based topology inference ideas may be useful. 
Second, most solutions to DMS problems involve either the inversion of the Laplacian matrix or an expensive optimization problem, yet they need to be performed at each decision opportunity. GSP tools of lower complexity could therefore be essential in this regard, and related theoretical results may prove to be useful in establishing error quantification bound for safe DMSs. 
Finally, an open challenge in the design of DMSs is the control over high-dimensional processes. From a network optimization perspective, several works have provided appropriate models to characterize systems evolving on high-dimensional networks \cite{Movric:2013,Zhang:2014,Salami:2017,Acemoglu:2011}). 
However, the theory and optimization techniques developed there do not directly apply to problems involving adaptive and online sequential decision strategies.
GSP tools could be adopted to overcome these limitations and tackle the challenge of online control of high-dimensional dynamic processes.}

\subsection{GSP and model interpretability}
\begin{figure}[t]
      \centering
        \subfloat[] {\includegraphics[width=7cm]{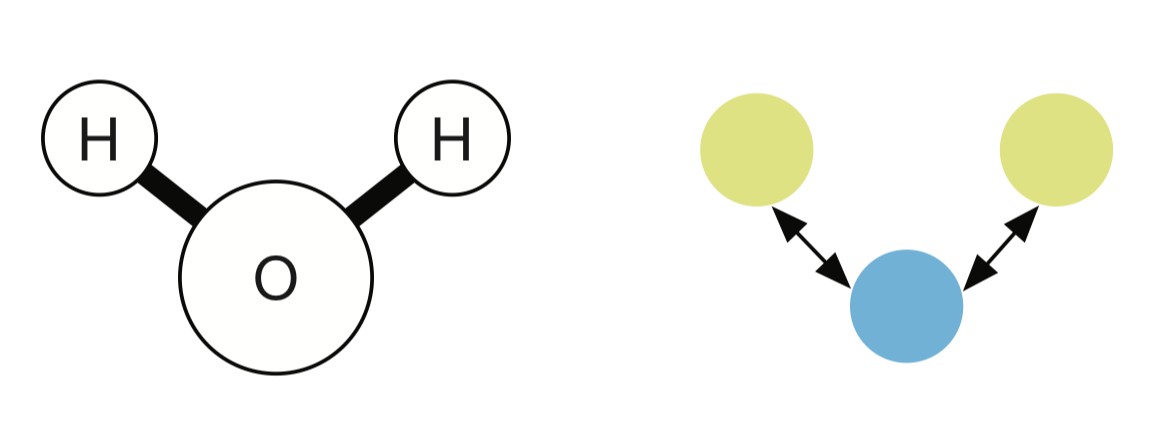}}~
        \subfloat[] {\includegraphics[width=7cm]{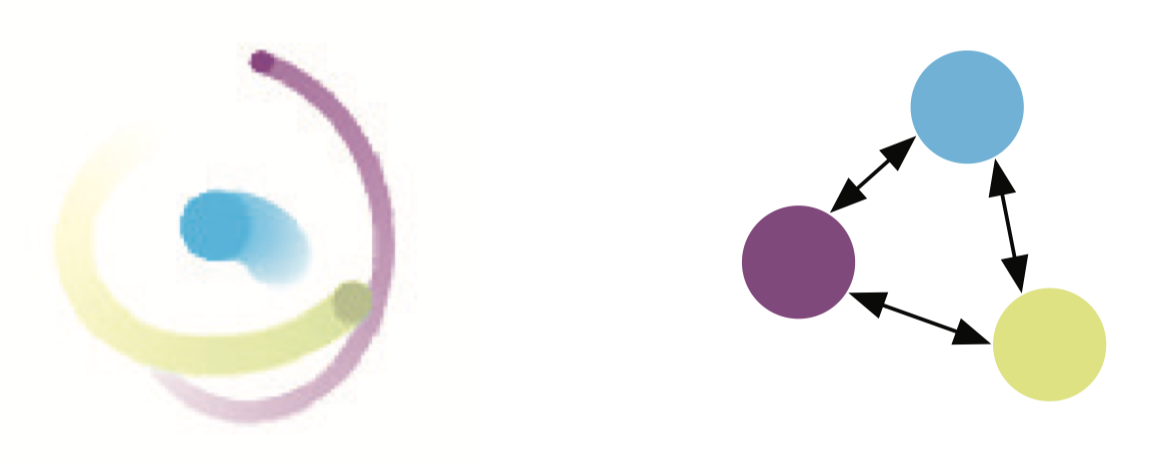}}
        \caption{Examples of relational structures in real-world systems. (a) A molecule (left) as a graph (right), where nodes represent atoms and edges correspond to bonds between them. (b) An n-body system (left) as a graph (right), where nodes represent bodies and edges correspond to connections between them. Figure from \cite{Battaglia_2018} with permission.}
        \label{fig:relational_structure}
\end{figure}

{GSP can play a significant role towards better defining what we call interpretable machine learning \cite{Gilpin18}. Currently, deep learning with large systems typically leads to non-robust and black-box decision systems, which is insufficient in safety critical applications such as personalized medicine and autonomous systems. Modeling the structure of the data with a graph could be a way of introducing domain knowledge (e.g., physical interactions such as the ones illustrated in Fig. \ref{fig:relational_structure}) in the learning process and eventually biases the system towards relational learning, which is a core ingredient for human-like intelligence \cite{Battaglia_2018}. For this reason, learning architectures that incorporate the graph structure, such as the GNNs described in Section \ref{sec:structure}, are significantly more explainable than typical black-box models.} By extending GNNs towards architectures that are built on more generic permutation invariant functions, we believe that a further step could be made towards relational learning. In addition, creating anisotropic filters \cite{Monti_2017} and adapting graph attention mechanisms such as the GAT could both open the door to more interpretable learning architectures.

\subsection{GSP and higher-order structures}
{There is a growing interest in recent years in analysing higher-order interactions in networks. In the complex systems literature, it has been recently shown that higher-order structures such as network motifs play a key role in detecting communities in complex networks \cite{benson2016higher}. Similar idea has been adopted in machine learning, where motifs has been used to design GNN models that are capable of handling directed graphs \cite{monti2018motifnet}. The extensions of GSP theories to deal with higher-order structures, such as motifs, simplicial complexes and hypergraphs, are an interesting direction, and the topological signal processing framework proposed in \cite{barbarossa2019topological} is a promising first step. Such extensions might reveal more complex structure within the data that goes beyond pairwise interactions; in addition, they may also help define machine learning models with improved interpretability.}

\bibliographystyle{IEEEbib.bst}
\bibliography{mybibfile.bib}

\begin{thebibliography}{10}

\bibitem{Zhu05}
X.~Zhu,
\newblock ``{Semi-supervised learning with graphs},''
\newblock {\em PhD thesis, Carnegie Mellon University, CMU-LTI-05-192}, 2005.

\bibitem{Fortunato10}
S.~Fortunato,
\newblock ``{Community detection in graphs},''
\newblock {\em Physics Reports}, vol. 486, no. 3-5, pp. 75--174, 2010.

\bibitem{Nickel16}
M.~Nickel, K.~Murphy, V.~Tresp, and E.~Gabrilovich,
\newblock ``A review of relational machine learning for knowledge graphs,''
\newblock {\em Proceedings of the IEEE}, vol. 104, no. 1, pp. 11--33, 2016.

\bibitem{Szklarczyk19}
D.~Szklarczyk et~al.,
\newblock ``String v11: Protein-protein association networks with increased
  coverage, supporting functional discovery in genome-wide experimental
  datasets,''
\newblock {\em Nucleic Acids Research}, vol. 47, no. 1, pp. 607--613, 2019.

\bibitem{Shuman13}
{D. I Shuman}, S.~K. Narang, P.~Frossard, A.~Ortega, and P.~Vandergheynst,
\newblock ``The emerging field of signal processing on graphs: Extending
  high-dimensional data analysis to networks and other irregular domains,''
\newblock {\em IEEE Signal Processing Magazine}, vol. 30, no. 3, pp. 83--98,
  2013.

\bibitem{Sandryhaila13}
A.~Sandryhaila and J.~M.~F. Moura,
\newblock ``{Discrete signal processing on graphs},''
\newblock {\em IEEE Transactions on Signal Processing}, vol. 61, no. 7, pp.
  1644--1656, 2013.

\bibitem{Ortega18}
A.~Ortega, P.~Frossard, J.~Kova{\v c}evi{\'c}, J.~M.~F. Moura, and
  P.~Vandergheynst,
\newblock ``{Graph signal processing: Overview, challenges and applications},''
\newblock {\em Proceedings of the IEEE}, vol. 106, no. 5, pp. 808--828, 2018.

\bibitem{bronstein2017geometric}
M.~M Bronstein, J.~Bruna, Y.~LeCun, A.~Szlam, and P.~Vandergheynst,
\newblock ``Geometric deep learning: Going beyond euclidean data,''
\newblock {\em IEEE Signal Processing Magazine}, vol. 34, no. 4, pp. 18--42,
  2017.

\bibitem{Wu19}
Z.~Wu, S.~Pan, F.~Chen, G.~Long, C.~Zhang, and P.~S. Yu,
\newblock ``A comprehensive survey on graph neural networks,''
\newblock {\em IEEE Transactions on Neural Networks and Learning Systems},
  2020.

\bibitem{Tremblay19}
N.~Tremblay and A.~Loukas,
\newblock ``Approximating spectral clustering via sampling: A review,''
\newblock in {\em Sampling Techniques for Supervised or Unsupervised Tasks},
  F.~Ros and S.~Guillaume, Eds., pp. 129--183. Springer, 2020.

\bibitem{Mateos19}
G.~Mateos, S.~Segarra, A.~G. Marques, and A.~Ribeiro,
\newblock ``Connecting the dots: Identifying network structure via graph signal
  processing,''
\newblock {\em IEEE Signal Processing Magazine}, vol. 36, no. 3, pp. 16--43,
  2019.

\bibitem{Dong19}
X.~Dong, D.~Thanou, M.~Rabbat, and P.~Frossard,
\newblock ``Learning graphs from data: A signal representation perspective,''
\newblock {\em IEEE Signal Processing Magazine}, vol. 36, no. 3, pp. 44--63,
  2019.

\bibitem{Smola03}
A.~Smola and R.~Kondor,
\newblock ``{Kernels and regularization on graphs},''
\newblock in {\em Annual Conference on Computational Learning Theory}, 2003,
  pp. 144--158.

\bibitem{Alvarez12}
M.~A. {\'A}lvarez, L.~Rosasco, and N.~D. Lawrence,
\newblock ``Kernels for vector-valued functions: A review,''
\newblock {\em Foundations and Trends in Machine Learning}, vol. 4, no. 3, pp.
  195--266, 2012.

\bibitem{Venkitaraman18}
A.~Venkitaraman, S.~Chatterjee, and P.~Handel,
\newblock ``Gaussian processes over graphs,''
\newblock {\em arXiv:1803.05776}, 2018.

\bibitem{Zhi20}
Y.-C. Zhi, N.~C. Cheng, and X.~Dong,
\newblock ``Gaussian processes on graphs via spectral kernel learning,''
\newblock {\em arXiv:2006.07361}, 2020.

\bibitem{Krizhevsky12}
A.~Krizhevsky, I.~Sutskever, and G.~Hinton,
\newblock ``{ImageNet classification with deep convolutional neural
  networks},''
\newblock in {\em Advances in Neural Information Processing Systems 25}, 2012,
  pp. 1097--1105.

\bibitem{Bruna14}
J.~Bruna, W.~Zaremba, A.~Szlam, and Y.~LeCun,
\newblock ``{Spectral networks and deep locally connected networks on
  graphs},''
\newblock in {\em International Conference on Learning Representations}, 2014.

\bibitem{defferrard2016convolutional}
M.~Defferrard, X.~Bresson, and P.~Vandergheynst,
\newblock ``Convolutional neural networks on graphs with fast localized
  spectral filtering,''
\newblock in {\em Advances in Neural Information Processing Systems 29}, 2016,
  pp. 3844--3852.

\bibitem{Chapelle06}
O.~Chapelle, B.~Sch{\"o}lkopf, and A.~Zien,
\newblock {\em {Semi-supervised learning}},
\newblock MIT Press, 2006.

\bibitem{Kipf17}
T.~N. Kipf and M.~Welling,
\newblock ``{Semi-supervised classification with graph convolutional
  networks},''
\newblock in {\em International Conference on Learning Representations}, 2017.

\bibitem{Fu20}
G.~Fu, Y.~Hou, J.~Zhang, K.~Ma, B.~F. Kamhoua, and J.~Cheng,
\newblock ``Understanding graph neural networks from graph signal denoising
  perspectives,''
\newblock {\em arXiv:2006.04386}, 2020.

\bibitem{Luxburg07}
U.~von Luxburg,
\newblock ``{A tutorial on spectral clustering},''
\newblock {\em Statistics and Computing}, vol. 17, no. 4, pp. {395--416}, 2007.

\bibitem{Lambiotte10}
R.~Lambiotte,
\newblock ``Multi-scale modularity in complex networks,''
\newblock in {\em International Symposium on Modeling and Optimization in
  Mobile, Ad Hoc and Wireless Networks}, 2010, pp. 546--553.

\bibitem{Newman06}
M.~E.~J. Newman,
\newblock ``Modularity and community structure in networks,''
\newblock {\em Proceedings of the National Academy of Sciences}, vol. 103, no.
  23, pp. 8577--8582, 2006.

\bibitem{Tremblay14}
N.~Tremblay and P.~Borgnat,
\newblock ``Graph wavelets for multiscale community mining,''
\newblock {\em IEEE Transactions on Signal Processing}, vol. 62, no. 20, pp.
  5227--5239, 2014.

\bibitem{Hammond11}
D.~K. Hammond, P.~Vandergheynst, and R.~Gribonval,
\newblock ``{Wavelets on graphs via spectral graph theory},''
\newblock {\em Applied and Computational Harmonic Analysis}, vol. 30, no. 2,
  pp. 129--150, 2011.

\bibitem{Tremblay16}
N.~Tremblay, G.~Puy, R.~Gribonval, and P.~Vandergheynst,
\newblock ``Compressive spectral clustering,''
\newblock in {\em International Conference on Machine Learning}, 2016, pp.
  1002--1011.

\bibitem{Kipf16}
T.~N. Kipf and M.~Welling,
\newblock ``Variational graph auto-encoders,''
\newblock in {\em NIPS Workshop on Bayesian Deep Learning}, 2016.

\bibitem{monti2017geometric}
F.~Monti, M.~M. Bronstein, and X.~Bresson,
\newblock ``{Geometric matrix completion with recurrent multi-graph neural
  networks},''
\newblock in {\em Advances in Neural Information Processing Systems 30}, 2017,
  pp. 3697--3707.

\bibitem{Ye_2019}
M.~Ye, V.~Stankovic, L.~Stankovic, and G.~Cheung,
\newblock ``Robust deep graph based learning for binary classification,''
\newblock {\em arXiv:1912.03321}, 2019.

\bibitem{Bontonou19}
M.~Bontonou, C.~Lassance, G.~B. Hacene, V.~Gripon, J.~Tang, and A.~Ortega,
\newblock ``Introducing graph smoothness loss for training deep learning
  architectures,''
\newblock in {\em IEEE Data Science Workshop}, 2019, pp. 225--228.

\bibitem{Lassance19}
C.~E. R.~K. Lassance, V.~Gripon, and A.~Ortega,
\newblock ``Laplacian power networks: Bounding indicator function smoothness
  for adversarial defense,''
\newblock {\em International Conference on Learning Representations}, 2019.

\bibitem{Shi_2019}
X.~Shi, L.~Salewski, M.~Schiegg, Z.~Akata, and M.~Welling,
\newblock ``Relational generalized few-shot learning,''
\newblock {\em arXiv:1907.09557}, 2019.

\bibitem{garcia2018fewshot}
V.~Garcia and J.~Bruna,
\newblock ``Few-shot learning with graph neural networks,''
\newblock in {\em International Conference on Learning Representations}, 2018.

\bibitem{Deutsch_2019}
S.~Deutsch, A.~L. Bertozzi, and S.~Soatto,
\newblock ``{Zero shot learning with the isoperimetric loss},''
\newblock in {\em AAAI Conference on Artificial Intelligence}, 2020.

\bibitem{Nassif20}
R.~Nassif, S.~Vlaski, C.~Richard, J.~Chen, and A.~H. Sayed,
\newblock ``Multitask learning over graphs: An approach for distributed,
  streaming machine mearning,''
\newblock {\em IEEE Signal Processing Magazine}, vol. 37, no. 3, pp. 14--25,
  2020.

\bibitem{levie2019transferability}
R.~Levie, M.~M Bronstein, and G.~Kutyniok,
\newblock ``Transferability of spectral graph convolutional neural networks,''
\newblock {\em arXiv:1907.12972}, 2019.

\bibitem{Gama19}
F.~Gama, J.~Bruna, and A.~Ribeiro,
\newblock ``Stability properties of graph neural networks,''
\newblock {\em arXiv:1905.04497}, 2019.

\bibitem{Kenlay20}
H.~Kenlay, D.~Thanou, and X.~Dong,
\newblock ``{On the stability of polynomial spectral graph filters},''
\newblock in {\em IEEE International Conference on Acoustics, Speech and Signal
  Processing}, 2020, pp. 5350--5354.

\bibitem{Isufi17}
E.~Isufi, A.~Loukas, A.~Simonetto, and G.~Leus,
\newblock ``Filtering random graph processes over random time-varying graphs,''
\newblock {\em IEEE Transactions on Signal Processing}, vol. 65, no. 16, pp.
  4406--4421, 2017.

\bibitem{Ceci18b}
E.~Ceci and S.~Barbarossa,
\newblock ``{Robust graph signal processing in the presence of uncertainties on
  graph topology},''
\newblock in {\em IEEE International Workshop on Signal Processing Advances in
  Wireless Communications}, 2018.

\bibitem{Segarra18}
S.~Segarra and A.~Ribeiro,
\newblock ``{Stability and continuity of centrality measures in weighted
  graphs},''
\newblock in {\em IEEE International Conference on Acoustics, Speech and Signal
  Processing}, 2018, pp. 3387--3391.

\bibitem{Sandryhaila14}
A.~Sandryhaila and J.~M.F. Moura,
\newblock ``Big data analysis with signal processing on graphs: Representation
  and processing of massive data sets with irregular structure,''
\newblock {\em IEEE Signal Processing Magazine}, vol. 31, no. 5, pp. 80--90,
  2014.

\bibitem{Banerjee08}
O.~Banerjee, L.~El Ghaoui, and A.~d'Aspremont,
\newblock ``Model selection through sparse maximum likelihood estimation for
  multivariate {G}aussian or binary data,''
\newblock {\em The Journal of Machine Learning Research}, vol. 9, pp. 485--516,
  2008.

\bibitem{Friedman08}
J.~Friedman, T.~Hastie, and R.~Tibshirani,
\newblock ``Sparse inverse covariance estimation with the graphical {L}asso,''
\newblock {\em Biostatistics}, vol. 9, no. 3, pp. 432--441, 2008.

\bibitem{GomezRodriguez_2010}
M.~Gomez-Rodriguez, J.~Leskovec, and A.~Krause,
\newblock ``Inferring networks of diffusion and influence,''
\newblock in {\em ACM SIGKDD International Conference on Knowledge Discovery
  and Data Mining}, 2010, pp. 1019--1028.

\bibitem{Myers2010}
S.~A. Myers and J.~Leskovec,
\newblock ``On the convexity of latent social network inference,''
\newblock in {\em Advances in Neural Information Processing Systems 23}, 2010,
  pp. 1741--1749.

\bibitem{Dong16}
X.~Dong, D.~Thanou, P.~Frossard, and P.~Vandergheynst,
\newblock ``Learning {L}aplacian matrix in smooth graph signal
  representations,''
\newblock {\em IEEE Transactions on Signal Processing}, vol. 64, no. 23, pp.
  6160--6173, 2016.

\bibitem{Kalofolias16}
V.~Kalofolias,
\newblock ``{How to learn a graph from smooth signals},''
\newblock in {\em International Conference on Artificial Intelligence and
  Statistics}, 2016, vol.~51, pp. 920--929.

\bibitem{Egilmez17}
H.~E. Egilmez, E.~Pavez, and A.~Ortega,
\newblock ``Graph learning from data under structural and {L}aplacian
  constraints,''
\newblock {\em IEEE Journal of Selected Topics in Signal Processing}, vol. 11,
  no. 6, pp. 825--841, 2017.

\bibitem{Chepuri17}
S.~P. Chepuri, S.~Liu, G.~Leus, and A.~O. Hero,
\newblock ``{Learning sparse graphs under smoothness prior},''
\newblock in {\em IEEE International Conference on Acoustics, Speech and Signal
  Processing}, 2017, pp. 6508--6512.

\bibitem{Thanou17}
D.~Thanou, X.~Dong, D.~Kressner, and P.~Frossard,
\newblock ``Learning heat diffusion graphs,''
\newblock {\em IEEE Transactions on Signal and Information Processing over
  Networks}, vol. 3, no. 3, pp. 484 -- 499, 2017.

\bibitem{Pasdeloup18}
B.~Pasdeloup, V.~Gripon, G.~Mercier, D.~Pastor, and M.~G. Rabbat,
\newblock ``Characterization and inference of graph diffusion processes from
  observations of stationary signals,''
\newblock {\em IEEE Transactions on Signal and Information Processing over
  Networks}, vol. 4, no. 3, pp. 481--496, 2018.

\bibitem{Segarra17a}
S.~Segarra, A.~G. Marques, G.~Mateos, and A.~Ribeiro,
\newblock ``Network topology inference from spectral templates,''
\newblock {\em IEEE Transactions on Signal and Information Processing over
  Networks}, vol. 3, no. 3, pp. 467--483, 2017.

\bibitem{Chen2020}
S.~Chen, C.~Duan, Y.~Yang, C.~Feng, D.~Li, and D.~Tian,
\newblock ``Deep unsupervised learning of {3D} point clouds via graph topology
  inference and filtering,''
\newblock {\em IEEE Transactions on Image Processing}, vol. 29, pp. 3183--3198,
  2020.

\bibitem{wang2019dynamic}
Y.~Wang, Y.~Sun, Z.~Liu, S.~E. Sarma, M.~M. Bronstein, and J.~M. Solomon,
\newblock ``Dynamic graph {CNN} for learning on point clouds,''
\newblock {\em ACM Transactions on Graphics}, vol. 38, no. 5, pp. 1--12, 2019.

\bibitem{Tong_2020}
A.~Tong, D.~van Dijk, J.~S. Stanley~III, M.~Amodio, K.~Yim, R.~Muhle,
  J.~Noonan, G.~Wolf, and S.~Krishnaswamy,
\newblock ``Interpretable neuron structuring with graph spectral
  regularization,''
\newblock {\em arXiv:1810.00424}, 2018.

\bibitem{Velickovic18}
P.~Veli{\v c}kovi{\'c}, G.~Cucurull, A.~Casanova, A.~Romero, P.~Li{\`o}, and
  Y.~Bengio,
\newblock ``Graph attention networks,''
\newblock in {\em International Conference on Learning Representations}, 2018.

\bibitem{Anirudh17}
R.~Anirudh, J.~J. Thiagarajan, R.~Sridhar, and T.~Bremer,
\newblock ``{MARGIN}: Uncovering deep neural networks using graph signal
  analysis,''
\newblock {\em arXiv:1711.05407}, 2017.

\bibitem{Gripon18}
V.~Gripon, A.~Ortega, and B.~Girault,
\newblock ``{An inside look at deep neural networks using graph signal
  processing},''
\newblock in {\em Information Theory and Applications Workshop}, 2018.

\bibitem{Miettinen19}
J.~Miettinen, S.~A. Vorobyov, and E.~Ollila,
\newblock ``Modelling graph errors: Towards robust graph signal processing,''
\newblock {\em arXiv:1903.08398}, 2019.

\bibitem{Zhang19}
Y.~Zhang, S.~Pal, M.~Coates, and D.~Üstebay,
\newblock ``{Bayesian graph convolutional neural networks for semi-supervised
  classification},''
\newblock in {\em AAAI Conference on Artificial Intelligence}, 2019, pp.
  5829--5836.

\bibitem{Mahadevan:2007}
S.~Mahadevan and M.~Maggioni,
\newblock ``Proto-value functions: A laplacian framework for learning
  representation and control in markov decision processes,''
\newblock {\em Journal of Machine Learning Research}, vol. 8, no. Oct, pp.
  2169--2231, 2007.

\bibitem{Toni:18}
L.~Toni and P.~Frossard,
\newblock ``Spectral {MAB} for unknown graph processes,''
\newblock in {\em European Signal Processing Conference}, 2018, pp. 116--120.

\bibitem{Movric:2013}
K.~H. Movric and F.~L. Lewis,
\newblock ``Cooperative optimal control for multi-agent systems on directed
  graph topologies,''
\newblock {\em IEEE Transactions on Automatic Control}, vol. 59, no. 3, pp.
  769--774, 2013.

\bibitem{Zhang:2014}
H.~Zhang, T.~Feng, G.-H. Yang, and H.~Liang,
\newblock ``Distributed cooperative optimal control for multiagent systems on
  directed graphs: An inverse optimal approach,''
\newblock {\em IEEE Transactions on Cybernetics}, vol. 45, no. 7, pp.
  1315--1326, 2014.

\bibitem{Salami:2017}
H.~Salami, B.~Ying, and A.~H Sayed,
\newblock ``Social learning over weakly connected graphs,''
\newblock {\em IEEE Transactions on Signal and Information Processing over
  Networks}, vol. 3, no. 2, pp. 222--238, 2017.

\bibitem{Acemoglu:2011}
D.~Acemoglu and A.~Ozdaglar,
\newblock ``Opinion dynamics and learning in social networks,''
\newblock {\em Dynamic Games and Applications}, vol. 1, no. 1, pp. 3--49, 2011.

\bibitem{Gilpin18}
L.~H. Gilpin, D.~Bau, B.~Z. Yuan, A.~Bajwa, M.~Specter, and L.~Kagal,
\newblock ``Compressive spectral clustering,''
\newblock in {\em IEEE International Conference on Data Science and Advanced
  Analytics}, 2018, pp. 80--89.

\bibitem{Battaglia_2018}
P.~W. Battaglia et~al.,
\newblock ``Relational inductive biases, deep learning, and graph networks,''
\newblock {\em arXiv:1806.01261}, 2018.

\bibitem{Monti_2017}
F.~Monti, D.~Boscaini, J.~Masci, E.~Rodola, J.~Svoboda, and M.~M. Bronstein,
\newblock ``Geometric deep learning on graphs and manifolds using mixture model
  {CNNs},''
\newblock in {\em {IEEE} Conference on Computer Vision and Pattern
  Recognition}, 2017, pp. 5115--5124.

\bibitem{benson2016higher}
A.~R Benson, D.~F Gleich, and J.~Leskovec,
\newblock ``Higher-order organization of complex networks,''
\newblock {\em Science}, vol. 353, no. 6295, pp. 163--166, 2016.

\bibitem{monti2018motifnet}
F.~Monti, K.~Otness, and M.~M Bronstein,
\newblock ``Motifnet: A motif-based graph convolutional network for directed
  graphs,''
\newblock in {\em IEEE Data Science Workshop}, 2018, pp. 225--228.

\bibitem{barbarossa2019topological}
S.~Barbarossa and S.~Sardellitti,
\newblock ``Topological signal processing over simplicial complexes,''
\newblock {\em arXiv:1907.11577}, 2019.

\end{thebibliography}

\end{document}